\pdfoutput=1

\documentclass[11pt]{article}

\usepackage[final]{ACL2023}

\usepackage{times}
\usepackage{latexsym}
\usepackage{graphicx} 
\usepackage[T1]{fontenc}

\usepackage[utf8]{inputenc}

\usepackage{microtype}
\usepackage{booktabs}
\usepackage{inconsolata}

\usepackage{amsmath}
\usepackage{amssymb}
\usepackage{titlesec}
\usepackage{caption}
\titlespacing{\section}{0pt}{3pt}{3pt} 
\titlespacing{\subsection}{0pt}{3pt}{3pt}
\title{Investigating the Impact of Data Selection Strategies on Language Model Performance}

\author{Jiayao Gu \\
  McGill University \\
  \texttt{\small jia.gu@mail.mcgill.ca} \\\And
  Liting Chen \\
  McGill University \\
  \texttt{\small liting.chen@mail.mcgill.ca}\\\And
  Yihong Li \\
  McGill University \\
  \texttt{\small yihong.li@mail.mcgill.ca}
}

\begin{document}

\maketitle
\begin{abstract}
Data selection is critical for enhancing the performance of language models, particularly when aligning training datasets with a desired target distribution. This study explores the effects of different data selection methods and feature types on model performance. We evaluate whether selecting data subsets can influence downstream tasks, whether n-gram features improve alignment with target distributions, and whether embedding-based neural features provide complementary benefits. Through comparative experiments using baseline random selection methods and distribution aligned approaches, we provide insights into the interplay between data selection strategies and model training efficacy. All code for this study can be found on \href{https://github.com/jgu13/HIR-Hybrid-Importance-Resampling-for-Language-Models}{github repository}.

\end{abstract}

\section{Introduction}

Large language models have garnered significant attention recently~\cite{bubeck2023sparks, liu2024visual, zhou2024large}. Their remarkable advancements can be attributed to the utilization of vast and continuously expanding text datasets for unsupervised pre-training~\citep{brown2020languagemodelsfewshotlearners}.
However, indiscriminately training a model on all accessible data may not yield the best results.
In our study, we study the problem of data selection~\citep{albalak2024surveydataselectionlanguage}. We examine the scenario where there is a large, varied raw dataset, characterized by distribution $q$, and a smaller dataset that represents a specific desired target distribution $p$. We aim to select a subset of the raw dataset that closely matches $p$.

Recent studies predominantly utilize heuristic methods to select training data. For instance, GPT-3~\citep{brown2020languagemodelsfewshotlearners} and PaLM~\citep{chowdhery2022palmscalinglanguagemodeling} employ a binary classifier $f(\cdot)$ to distinguish high-quality formal text $p$ from noisy web data $q$, selecting web examples that exhibit a high predicted probability $f(\boldsymbol{x})$.
However, as each example is considered independently, this method does not allow for the selection of a predetermined number of examples in a single pass. Moreover, such a sampling strategy tends to favor the prominent modes of the target distribution, thereby reducing the diversity of the samples.
The recent DSIR approach~\citep{xie2023data} introduces the use of hashed n-gram features to approximate the target distribution $\hat{p}$ and the raw distribution $\hat{q}$, and calculates the importance weights as $\frac{\hat{p}(\boldsymbol{x})}{\hat{q}(\boldsymbol{x})}$.
These importance weights are subsequently utilized for resampling. Although this method aligns the raw distribution with the target distribution to some extent, it relies solely on  n-gram features, which lack contextual awareness.

\begin{figure}[htbp]
    \centering
    \includegraphics[width=0.5\textwidth]{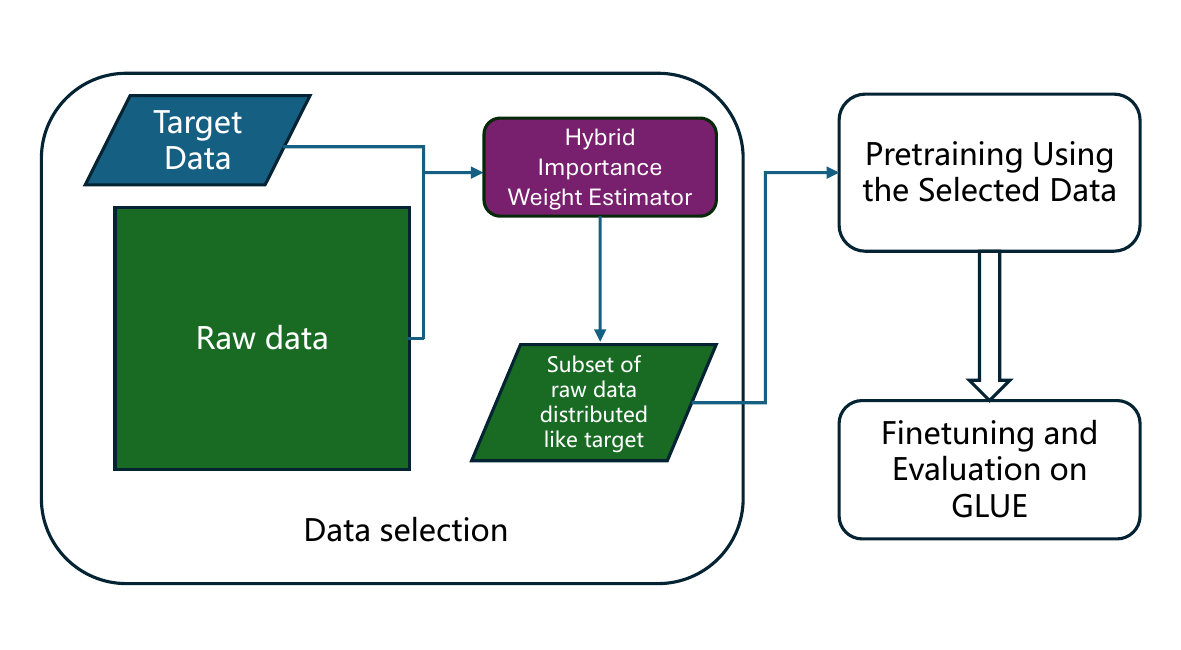} 
    \caption{ Data selection and pretraining pipeline. The process begins with target and raw datasets, where a subset of raw data is selected to match the target distribution. The selected data is used for pretraining, followed by fine-tuning and evaluation on GLUE tasks}
    \label{fig:importance_resampling}
\end{figure}

In this work, we  \textbf{investigate  the impact of data selection strategies on language model performance}. We implement random selection and DSIR as the baseline methods. We also propose \textit{HIR}, Hybrid Importance Resampling, where we combines statistical n-gram features and neural features to see if aligning raw datasets with a target using neural features can improve the performance of downstream tasks.
Specifically, for a dataset resembling the target set, we estimate two distinct distributions: an n-gram-based distribution, denoted as $\hat{p}_{\text{ngram}}(\boldsymbol{x})$, parameterized by a Multinomial Distribution, and a neural feature-based distribution, denoted as $\hat{p}_{\text{nn}}(\boldsymbol{x})$, parameterized by Gaussian Mixture Models. To balance these two perspectives, we introduce a weighting factor $\alpha$ and construct a hybrid distribution: $
\hat{p}_{\text{hybrid}}(\boldsymbol{x}) = \hat{p}_{\text{ngram}}(\boldsymbol{x})^{\alpha} \cdot \hat{p}_{\text{nn}}(\boldsymbol{x})^{1-\alpha},
$ which integrates the discrete properties of n-gram features with the continuous characteristics of neural features. Using the hybrid distribution, we compute the sample importance weight as $
\frac{\hat{p}_{\text{hybrid}}(\boldsymbol{x})}{\hat{q}_{\text{hybrid}}(\boldsymbol{x})},
$
where $\hat{q}_{\text{hybrid}}(\boldsymbol{x})$ represents the hybrid distribution of the raw dataset. Based on these importance weights, we resample a subset of $k$ samples from the raw dataset, assigning selection probabilities proportional to$
\frac{\hat{p}_{\text{hybrid}}(\boldsymbol{x})}{\hat{q}_{\text{hybrid}}(\boldsymbol{x})}.
$ The selected data is drawn from the Pile dataset and used for continued pretraining of language models.  Finally, we evaluate the performance of these models by fine-tuning on the GLUE benchmark~\cite{wang-etal-2018-glue}. The whole process is visualized in Figure~\ref{fig:importance_resampling}.

Our experimental findings highlight the critical role of data selection in improving language model performance. By aligning the target distribution with the raw dataset, we demonstrate  performance gains across multiple downstream tasks. The results confirm that carefully curating pretraining data to resemble the target distribution enhances the model's ability to generalize effectively.
Furthermore, the proposed hybrid importance reweighting (HIR) method with weight parameter equals to 0, which leverages neural embedding-based features, consistently improves over random selection. This improvement highlights its ability to align with the target dataset, capturing semantic and syntactic richness at a macro level. However, DSIR's superiority in most tasks underscores the value of token-level modeling with n-gram statistics for pretraining language model. 


\section{Related work}

The selection of pretraining data has been widely studied in language modeling. Prior methods focus on heuristic approaches, manual data curation, and distributional matching \cite{brown2020languagemodelsfewshotlearners, chowdhery2022palmscalinglanguagemodeling, zhang2024pretrainingdatadetectionlarge, xie2023data}. This section reviews these strategies, identifying their strengths and limitations in the context of large-scale data selection.

\textbf{Heuristic selection} filters data using a simple binary classifier or predefined rules to iteratively determine whether an instance belongs to the target set or not. It cannot guarantee that selected data is distributed like formal text \cite{brown2020languagemodelsfewshotlearners}. It may include informal or noisy content, and focus on surface-level features like word overlaps, which may miss deeper semantic or contextual relevance.

\textbf{Manual data curation} relies on manually curated, pre-existing categorized datasets \cite{gururangan2020dontstoppretrainingadapt, zhang2024pretrainingdatadetectionlarge}. These datasets are shaped by the criteria and decisions of their original creators, potentially embedding biases and omitting valuable content. Additionally, pre-curated datasets are static, and their broad categorization may lack the granularity needed for specialized tasks. These limitations highlight the need for more dynamic and automated methods to refine domain-specific data selection.

\textbf{Distributional Matching:} The DSIR method relies on hashed n-gram features for efficient data selection \cite{xie2023data}; as a result, it captures word-level overlaps, losing abstract semantic and syntactic information crucial for matching complex target distributions. Moreover, it does not optimize the importance of specific features, potentially overlooking those most relevant for downstream tasks. Hashing also introduces noise due to collisions, reducing precision in representing the data. While efficient and scalable, these limitations highlight the need for other feature extraction and optimization to improve alignment with complex target distributions.
Although bi-level optimization methods exist for reweighting training data to align with the validation set~\cite{chen2022gradient, chen2021generalized, ren2018learning}, they are typically computationally expensive due to the involvement of second-order gradients.

\section{Data Selection}

Given a small number of target set $X_1, X_2, \dots, X_n$ 
from a target distribution $p$ and a raw dataset 
$Y_1, Y_2, \dots, Y_m$ from a distribution $q$ with $n \ll m$, the objective is to extract a subset from the raw dataset of size $k$ that closely aligns with the information in the target set. 

We begin by introducing two baseline methods for data selection: random selection and DSIR. Random selection provides a straightforward benchmark, while DSIR leverages unigram and bigram statistics to capture token-level features. We then present HIR method, which integrates both n-gram and neural embedding-based features to construct a hybrid distribution.

Heuristic methods were not included in this study because ensuring a consistent subset size during selection is challenging with heuristic approaches. Without matching subset sizes, it becomes difficult to make fair comparisons, as differences in subset size could bias the evaluation of the methods. Therefore, we focused on methods that guarantee subsets of equal size for a more rigorous and fair analysis.

\subsection{Baseline Pretraining Dataset Selection}

\textbf{Random selection: } $k$ indices are randomly chosen from the range $1$ to $m$. The data corresponding to the selected indices is then extracted and aligned for further use.

\textbf{DSIR:} This method is designed to select a subset of data from a large raw dataset such that it closely matches the distribution of a smaller target dataset \cite{xie2023data}. It represents each sample in the raw dataset in a reduced feature vector and estimates its importance weight, which is subsequently used to re-sample samples from the raw dataset. 

\textbf{Hashed N-gram features } Each sample $x$ is represented in an m-dimensional feature vector $z\in\mathbb{N}^m$ $(m=10000)$ using hashed n-gram features, where $z$ contains the counts of unigrams and bigrams for sample $x$. For example, if $x$ is "I love Montreal.", the list of unigrams and bigrams is \texttt{[I, love, Montreal, I love, love Montreal]}. Hash each element of this list to get the list indices \texttt{ [0, 2, 3, 4, 4]}. The counts of each index are \texttt{ [1, 0, 1, 1, 2, ...,0]}. 

\textbf{Bag of hashed n-gram model } We use $q_{\text{ng}}$ and $p_{\text{ng}}$ to denote raw and target dataset distributions respectively. $q_{\text{ng}}$ or $p_{\text{ng}}$ is parameterized using bag of hashed n-gram models. In this model, the probability of a feature vector $z\in\mathbb{N}^m$ is $P(z; \gamma) = \prod_{j=1}^m\gamma[j]^{z[j]}$. $\gamma$ is an m-dimensional probability vector where each probability is estimated using feature vectors $z_1,\dots,z_s$: $\gamma=\frac{1}{\sum_{j=1}^s\sum_{i=1}^m z_{ji}}\sum_{j=1}^sz_j$.



\subsection{Hybrid Importance Resampling } Although DSIR aligns the raw distribution with the target distribution on local context, it does not take into account the global context in which the sample is located. Therefore, we proposed to incorporate the neural network features to compensate for hashed n-gram features to represent both the local and global context of a sample. 

\textbf{Fit Gaussian Mixture Model to Neural Network Features } Each sample \( x \) is embedded in a 384-dimensional feature space using SentenceTransformer \cite{DBLP:journals/corr/abs-1908-10084} (\texttt{all-MiniLM-L6-v2}). We define \( q_{\text{nn}} \) and \( p_{\text{nn}} \) as the distributions of the raw and target datasets, respectively. \( q_{\text{nn}} \) is estimated using a diagonal Gaussian Mixture Model (GMM) with 1000 components, while \( p_{\text{nn}} \) is estimated with 50 components. To manage memory efficiently, the GMMs for \( q_{\text{nn}} \) and \( p_{\text{nn}} \) are fitted iteratively on dataset chunks. Each chunk is loaded into memory to fit an initial GMM, which is then refined iteratively using subsequent chunks. The weights, means, and covariances from the previous iteration serve as the starting point for the next.

\textbf{Hybrid importance weights } We determine a weight parameter $\alpha$ to balance the distributions $p_{\text{ng}}$ and $p_{\text{nn}}$, forming a hybrid distribution: $p_{\text{hybrid}}(x) = p_{\text{ng}}(x) ^ {\alpha} \cdot p_{\text{nn}}(x)^{1 - \alpha}$, which captures both the local characteristics of hashed n-gram features as well as the global neural network features. Subsequently, we compute the sample importance weights as $\omega_i^{\text{hybrid}} = \frac{p_{\text{hybrid}}(x)}{q_{\text{hybrid}}(x)}$. The computation can be simplified further to $\omega_i^{\text{hybrid}} = {\omega^{\text{ng}}_i}^\alpha \cdot {\omega_i^{nn}}^{(1 - \alpha)}$ where $\omega_i^{\text{ng}} = \frac{p_{\text{ng}}(x)}{q_{\text{ng}}(x)}$ and $\omega_i^{\text{nn}} = \frac{p_{\text{nn}}(x)}{q_{\text{nn}}(x)}$.

\textbf{Resampling}
We sample $k$ samples without replacement with probabilities $\omega_i^{\text{hybrid}}$. When $\alpha = 1$, $\omega_i^{\text{hybrid}} = \omega_i^{\text{ng}}$, and when $\alpha = 0$, $\omega_i^{\text{hybrid}} = \omega_i^{\text{nn}}$. In this report, we focus on $\alpha = 0$, which corresponds to using the neural embedding-based distribution for data selection.  The decision to test only $\alpha = 0$ stems from the complexity involved in fine-tuning this parameter. Exploring the optimal value of $\alpha$ would require extensive experimentation across a wide range of tasks and datasets, which is beyond the scope of this study. By setting $\alpha = 0$, we aim to evaluate the raw effectiveness of the GMM-based approach without additional confounding factors. The selection of an optimal $\alpha$ value, which could balance n-gram and neural embedding contributions, is left as an avenue for future work. To select samples, the raw dataset was partitioned and sorted such that the $k^{\text{th}}$ sample occupied the final sorted position. The top-$k$ samples, as determined by their $\omega_i^{\text{hybrid}}$ probabilities, were then selected.  

\section{Pretraining}
In the pretraining phase, the text is tokenized into subword units using a tokenizer. A data collator is then initialized to randomly mask 15\% of the tokens in the input text. For the masked tokens, 80\% of the time they are replaced with the `[MASK]` token, 10\% of the time with a random word, and 10\% of the time they are left unchanged. The model is trained on this masked data to predict the original tokens for the masked positions based on their surrounding context, enabling it to learn contextual relationships in the text. The loss is computed by first generating the probability distribution over the vocabulary for each predicted word using the model's output logits. Then, the cross-entropy loss is calculated between this predicted probability distribution and the true labels for the masked positions, which guides the model to improve its predictions during training.
\section{ Finetuning and Evaluation}
Six tasks from the GLUE benchmarks are used to fine-tune and evaluate the models: these tasks were selected to provide a comprehensive evaluation of the model's performance across various natural language understanding challenges. Due to the significant time required for training on some tasks in GLUE, we strategically limited our selection to six tasks to balance computational efficiency with evaluation depth. 
\paragraph{COLA (Corpus of Linguistic Acceptability)}
COLA focuses on checking if a sentence is grammatically acceptable. It uses examples labeled by linguists as either acceptable or unacceptable. The main metric is the Matthews Correlation Coefficient (MCC), which measures how well the model's predictions align with the correct labels.

\paragraph{MRPC (Microsoft Research Paraphrase Corpus)}
MRPC is about determining if two sentences mean the same thing. The dataset has pairs of sentences, often from news, labeled as either semantically equivalent or not. The performance is measured using Accuracy (how many predictions are correct).

\paragraph{QNLI (Question Natural Language Inference)}
QNLI involves deciding whether a sentence correctly answers a question. The main metric used is Accuracy.

\paragraph{RTE (Recognizing Textual Entailment)}
RTE is about understanding whether a sentence (the hypothesis) logically follows from another sentence (the premise). The dataset contains pairs of sentences labeled as entailment or not. Accuracy is the metric used to evaluate performance.

\paragraph{SST-2 (Stanford Sentiment Treebank - Binary Classification)}
SST-2 is a task that involves identifying whether a sentence expresses a positive or negative sentiment. The dataset comes from movie reviews labeled as positive or negative. The metric for this task is Accuracy, which measures the percentage of correct classifications.

\paragraph{STS-B (Semantic Textual Similarity Benchmark)}
STS-B is about measuring how similar two sentences are in meaning. Each pair of sentences is scored on a scale from 0 (no similarity) to 5 (identical in meaning). The metric we used is Spearman Correlation, which compares the ranking order of the predictions with the true scores.
We conclude all of the tasks and the metrics we used in Table \ref{tb:1}.
\begin{table}[h]
\centering
\begin{tabular}{@{}ll@{}}
\toprule
\textbf{Task} & \textbf{Metric} \\ \midrule
COLA & MCC \\
MRPC & Accuracy \\
QNLI & Accuracy \\
RTE & Accuracy \\
SST-2 & Accuracy \\
STS-B & Spearman Correlation \\ \bottomrule
\end{tabular}
\caption{Metrics used for each task.}
\label{tb:1}
\end{table}


\section{Experiments and Results}

In the experiment, we aimed to use the Pile dataset as our raw dataset. 
When selecting data , we found that the original Pile dataset had been taken down due to copyright concerns. As an alternative, we used the uncopyrighted version of the Pile dataset available on \href{https://huggingface.co/datasets/monology/pile-uncopyrighted}{Hugging Face}, which excludes copyrighted subsets such as \textit{Books3}, \textit{BookCorpus2}, \textit{OpenSubtitles}, \textit{YTSubtitles}, and \textit{OWT2}. Moreover, due to limited computational resources, we were unable to perform data selection on the entire 800 GB dataset. Instead, without loss of generality, we conducted the experiment on a proportionally smaller raw dataset and target set. Consequently, the resulting selected subset was also proportionally smaller. 

For the baseline data selection methods we evaluated, we used the first 58 million documents of the uncopyrighted Pile dataset as the raw dataset and the subsets labeled \textit{Gutenberg (PG-19)} and \textit{Wikipedia (en)} as the target set. These two target sets provide a balanced combination of high-quality, diverse, and large-scale data. This enables the model to generalize well to various natural language understanding tasks in GLUE. Our objective was to select 1.7 million instances from the raw dataset that closely matched the distribution of the target dataset. However, for the HIR selection method, due to computational limitations, we only selected 47,000 instances from the first 1.4 million documents of the raw dataset. To ensure consistency in dataset size, we replaced the first 47,000 instances selected by DSIR with the 47,000 instances selected by HIR.

We summarize our findings in Table~\ref{tab:results}. For each task, we adopted the hyperparameter settings described in~\cite{xie2023data} and conducted experiments using three random seeds. 

\begin{table}[h!]
\centering
\begin{tabular}{@{}lccc@{}}
\toprule
\textbf{Task} & \textbf{Random } & \textbf{DSIR} & \textbf{HIR}\\
\midrule
COLA  & $18.73_{10.22}$ & $\textbf{28.42}_{10.59}$ & $25.94_{8.96}$ \\
MRPC  & $82.68_{0.61}$  & $\textbf{82.84}_{1.72}$  & $79.90_{1.49}$ \\
QNLI  & $85.32_{0.33}$  & $\textbf{85.76}_{0.67}$  & $85.23_{0.14}$ \\
RTE   & $56.92_{1.50}$  & $\textbf{61.25}_{0.91}$  & $60.05_{1.46}$ \\
SST2  & $85.01_{0.35}$  & $\textbf{87.35}_{0.99}$  & $86.85_{0.86}$ \\
STSB  & $85.06_{0.12}$  & $85.20_{0.05}$  & $\textbf{85.71}_{0.18}$ \\
\bottomrule
\end{tabular}
\caption{Performance comparison across tasks, reporting mean and standard deviation (subscript). Best performance is bolded.}
\label{tab:results}
\end{table}
\textbf{Impact of Data Selection}
By comparing the results of random selection and the two other methods, we can conclude that models pretrained on datasets selected through n-gram and embedding-based methods exhibit superior alignment with target data distributions. This improved alignment leads to consistent performance gains on GLUE tasks compared to random selection.

\textbf{Utility of Neural Embeddings}
 HIR demonstrates superior performance by capturing semantic and syntactic richness. These neural features are effective in aligning the raw dataset with the target distribution, enabling improved pretraining outcomes. While DSIR excels in capturing token-level characteristics, HIR’s ability to align distributions at a broader contextual level highlights its value as a complementary approach. 

\textbf{Effectiveness of N-Gram Features}
DSIR consistently demonstrates superior performance, achieving the best results in five out of the six tasks (COLA, MRPC, QNLI, RTE, and SST2). This highlights its effectiveness in capturing the token-level characteristics essential for improving language model pretraining.

\section{Discussion and Limitation}
By estimating sample importance weights based primarily on uni-grams and bi-grams, DSIR excels at capturing local token-level features that align closely with the objectives of language model training. In contrast, GMM—which uses sentence-level embeddings—focuses on higher-level, contextual features between sentences. While this broader semantic perspective can be useful in certain scenarios, it is less directly aligned with the low-level token prediction tasks that most language models on, such as predicting masked tokens based on local context. As a result, DSIR’s emphasis on n-gram statistics is generally more effective for guiding sample selection in language model pretraining. 

Additionally, applying a diagonal GMM to high-dimensional embedding spaces may not yield accurate probability estimates of the raw and target distributions. If these estimates are imprecise, the computed importance weights will be less reliable, potentially degrading language model training quality. Moreover, if the target distribution does not markedly differ from the raw distribution at the sentence-level scale, then GMM-based methods will struggle to capture useful differences. This lack of meaningful divergence at the global context level reduces GMM’s ability to identify informative samples, further explaining why GMM-based importance resampling may underperform DSIR.

Despite the promising results, our analysis has some limitations. First, we conducted experiments on a relatively small subset of the Pile dataset and trained the models using the data only once. This approach might not generalize well to larger-scale datasets or iterative training scenarios. Second, we did not perform extensive hyperparameter tuning, particularly for the parameter $\alpha$ in the importance weights calculations, which could potentially influence the effectiveness of the data selection methods. Additionally, the computational overhead of calculating importance weights, especially for neural feature-based methods, was not addressed in detail. This could be a limitation when scaling to  large datasets or high-dimensional feature spaces.

\section{Conclusion}
In conclusion, we investigated data selection methods by incorporating neural features in computing the importance weights of raw dataset. We found data selection improves performance of language models by comparing randomly selected data with n-gram and neural feature-based methods. Neural based method surpasses n-gram-based method on capturing semantic and syntactic richness, while n-gram-based method consistently excels in capturing token-level features. Overall, n-gram-based method outperforms neural based model on most of the tasks from the GLUE benchmarks. Future work could explore the impact of and strategies for selecting the weight parameter.
\section{Statement of Contributions}
This project was a collaborative effort among all team members. Below, we outline the specific contributions of each member:
\vspace{-0.5em}
\begin{itemize}
    \item \textbf{Jiayao Gu}: Responsible for the implementation of the HIR method for data selection, combining n-gram and neural features to align raw datasets with the target distribution, and resampling.
    \vspace{-0.5em}
    \item \textbf{Yihong Li}: Focused on the pretraining stage, ensuring that the model was prepared for downstream tasks. Data collection and pre-processing.
    \vspace{-0.5em}
    \item \textbf{Liting Chen}: Handled fine-tuning and evaluation using GLUE tasks, ensuring comprehensive testing and performance assessment.
\end{itemize}
\vspace{-0.5em}
In addition to these specific tasks in implementation, all members participated in the design of the project and contributed to writing.
We would like to thank \textbf{Can Chen} from Mila for providing the idea for the project.

\section{Ethics Statement}
This work explores the development of improved data selection methods for pretraining language models, specifically targeting alignment between raw datasets and desired distributions. Our research is conducted using publicly available datasets, including the Pile, Gutenberg (PG-19), and Wikipedia (en). These datasets are selected based on their established use in the research community and their relevance to the task.

We ensure compliance with the terms of use and licensing agreements associated with the datasets and limit our scope to improving data selection methodologies. This work is intended for academic and research purposes, with the goal of advancing techniques for language model training efficiency and quality. We encourage the responsible application of our methods in accordance with community standards and research best practices.

\bibliography{custom}

\begin{thebibliography}{14}
\expandafter\ifx\csname natexlab\endcsname\relax\def\natexlab#1{#1}\fi

\bibitem[{Albalak et~al.(2024)Albalak, Elazar, Xie, Longpre, Lambert, Wang, Muennighoff, Hou, Pan, Jeong, Raffel, Chang, Hashimoto, and Wang}]{albalak2024surveydataselectionlanguage}
Alon Albalak, Yanai Elazar, Sang~Michael Xie, Shayne Longpre, Nathan Lambert, Xinyi Wang, Niklas Muennighoff, Bairu Hou, Liangming Pan, Haewon Jeong, Colin Raffel, Shiyu Chang, Tatsunori Hashimoto, and William~Yang Wang. 2024.
\newblock \href {http://arxiv.org/abs/2402.16827} {A survey on data selection for language models}.

\bibitem[{Brown et~al.(2020)Brown, Mann, Ryder, Subbiah, Kaplan, Dhariwal, Neelakantan, Shyam, Sastry, Askell, Agarwal, Herbert-Voss, Krueger, Henighan, Child, Ramesh, Ziegler, Wu, Winter, Hesse, Chen, Sigler, Litwin, Gray, Chess, Clark, Berner, McCandlish, Radford, Sutskever, and Amodei}]{brown2020languagemodelsfewshotlearners}
Tom~B. Brown, Benjamin Mann, Nick Ryder, Melanie Subbiah, Jared Kaplan, Prafulla Dhariwal, Arvind Neelakantan, Pranav Shyam, Girish Sastry, Amanda Askell, Sandhini Agarwal, Ariel Herbert-Voss, Gretchen Krueger, Tom Henighan, Rewon Child, Aditya Ramesh, Daniel~M. Ziegler, Jeffrey Wu, Clemens Winter, Christopher Hesse, Mark Chen, Eric Sigler, Mateusz Litwin, Scott Gray, Benjamin Chess, Jack Clark, Christopher Berner, Sam McCandlish, Alec Radford, Ilya Sutskever, and Dario Amodei. 2020.
\newblock \href {http://arxiv.org/abs/2005.14165} {Language models are few-shot learners}.

\bibitem[{Bubeck et~al.(2023)Bubeck, Chandrasekaran, Eldan, Gehrke, Horvitz, Kamar, Lee, Lee, Li, Lundberg et~al.}]{bubeck2023sparks}
Sbastien Bubeck, Varun Chandrasekaran, Ronen Eldan, Johannes Gehrke, Eric Horvitz, Ece Kamar, Peter Lee, Yin~Tat Lee, Yuanzhi Li, Scott Lundberg, et~al. 2023.
\newblock Sparks of artificial general intelligence: Early experiments with gpt-4.
\newblock \emph{arXiv preprint arXiv:2303.12712}.

\bibitem[{Chen et~al.(2022)Chen, Chen, Ma, Liu, and Liu}]{chen2022gradient}
Can Chen, Xi~Chen, Chen Ma, Zixuan Liu, and Xue Liu. 2022.
\newblock Gradient-based bi-level optimization for deep learning: A survey.
\newblock \emph{arXiv preprint arXiv:2207.11719}.

\bibitem[{Chen et~al.(2021)Chen, Zheng, Chen, Dong, Liu, Liu, and Dou}]{chen2021generalized}
Can Chen, Shuhao Zheng, Xi~Chen, Erqun Dong, Xue~Steve Liu, Hao Liu, and Dejing Dou. 2021.
\newblock Generalized dataweighting via class-level gradient manipulation.
\newblock \emph{Advances in Neural Information Processing Systems}.

\bibitem[{Chowdhery et~al.(2022)Chowdhery, Narang, Devlin, Bosma, Mishra, Roberts, Barham, Chung, Sutton, Gehrmann, Schuh, Shi, Tsvyashchenko, Maynez, Rao, Barnes, Tay, Shazeer, Prabhakaran, Reif, Du, Hutchinson, Pope, Bradbury, Austin, Isard, Gur-Ari, Yin, Duke, Levskaya, Ghemawat, Dev, Michalewski, Garcia, Misra, Robinson, Fedus, Zhou, Ippolito, Luan, Lim, Zoph, Spiridonov, Sepassi, Dohan, Agrawal, Omernick, Dai, Pillai, Pellat, Lewkowycz, Moreira, Child, Polozov, Lee, Zhou, Wang, Saeta, Diaz, Firat, Catasta, Wei, Meier-Hellstern, Eck, Dean, Petrov, and Fiedel}]{chowdhery2022palmscalinglanguagemodeling}
Aakanksha Chowdhery, Sharan Narang, Jacob Devlin, Maarten Bosma, Gaurav Mishra, Adam Roberts, Paul Barham, Hyung~Won Chung, Charles Sutton, Sebastian Gehrmann, Parker Schuh, Kensen Shi, Sasha Tsvyashchenko, Joshua Maynez, Abhishek Rao, Parker Barnes, Yi~Tay, Noam Shazeer, Vinodkumar Prabhakaran, Emily Reif, Nan Du, Ben Hutchinson, Reiner Pope, James Bradbury, Jacob Austin, Michael Isard, Guy Gur-Ari, Pengcheng Yin, Toju Duke, Anselm Levskaya, Sanjay Ghemawat, Sunipa Dev, Henryk Michalewski, Xavier Garcia, Vedant Misra, Kevin Robinson, Liam Fedus, Denny Zhou, Daphne Ippolito, David Luan, Hyeontaek Lim, Barret Zoph, Alexander Spiridonov, Ryan Sepassi, David Dohan, Shivani Agrawal, Mark Omernick, Andrew~M. Dai, Thanumalayan~Sankaranarayana Pillai, Marie Pellat, Aitor Lewkowycz, Erica Moreira, Rewon Child, Oleksandr Polozov, Katherine Lee, Zongwei Zhou, Xuezhi Wang, Brennan Saeta, Mark Diaz, Orhan Firat, Michele Catasta, Jason Wei, Kathy Meier-Hellstern, Douglas Eck, Jeff Dean, Slav Petrov, and Noah Fiedel. 2022.
\newblock \href {http://arxiv.org/abs/2204.02311} {Palm: Scaling language modeling with pathways}.

\bibitem[{Gururangan et~al.(2020)Gururangan, Marasović, Swayamdipta, Lo, Beltagy, Downey, and Smith}]{gururangan2020dontstoppretrainingadapt}
Suchin Gururangan, Ana Marasović, Swabha Swayamdipta, Kyle Lo, Iz~Beltagy, Doug Downey, and Noah~A. Smith. 2020.
\newblock \href {http://arxiv.org/abs/2004.10964} {Don't stop pretraining: Adapt language models to domains and tasks}.

\bibitem[{Liu et~al.(2024)Liu, Li, Wu, and Lee}]{liu2024visual}
Haotian Liu, Chunyuan Li, Qingyang Wu, and Yong~Jae Lee. 2024.
\newblock Visual instruction tuning.
\newblock \emph{Advances in neural information processing systems}, 36.

\bibitem[{Reimers and Gurevych(2019)}]{DBLP:journals/corr/abs-1908-10084}
Nils Reimers and Iryna Gurevych. 2019.
\newblock \href {http://arxiv.org/abs/1908.10084} {Sentence-bert: Sentence embeddings using siamese bert-networks}.
\newblock \emph{CoRR}, abs/1908.10084.

\bibitem[{Ren et~al.(2018)Ren, Zeng, Yang, and Urtasun}]{ren2018learning}
Mengye Ren, Wenyuan Zeng, Bin Yang, and Raquel Urtasun. 2018.
\newblock Learning to reweight examples for robust deep learning.
\newblock In \emph{International conference on machine learning}. PMLR.

\bibitem[{Wang(2018)}]{wang-etal-2018-glue}
Alex Wang. 2018.
\newblock Glue: A multi-task benchmark and analysis platform for natural language understanding.
\newblock \emph{arXiv preprint arXiv:1804.07461}.

\bibitem[{Xie et~al.(2023)Xie, Santurkar, Ma, and Liang}]{xie2023data}
Sang~Michael Xie, Shibani Santurkar, Tengyu Ma, and Percy~S Liang. 2023.
\newblock Data selection for language models via importance resampling.
\newblock \emph{Advances in Neural Information Processing Systems}.

\bibitem[{Zhang et~al.(2024)Zhang, Zhang, Guo, de~Rijke, Fan, and Cheng}]{zhang2024pretrainingdatadetectionlarge}
Weichao Zhang, Ruqing Zhang, Jiafeng Guo, Maarten de~Rijke, Yixing Fan, and Xueqi Cheng. 2024.
\newblock \href {http://arxiv.org/abs/2409.14781} {Pretraining data detection for large language models: A divergence-based calibration method}.

\bibitem[{Zhou et~al.(2024)Zhou, Hu, Yuan, Cui, Jin, Chen, Wu, Yuan, Jiang, Wu et~al.}]{zhou2024large}
Hao Zhou, Chengming Hu, Ye~Yuan, Yufei Cui, Yili Jin, Can Chen, Haolun Wu, Dun Yuan, Li~Jiang, Di~Wu, et~al. 2024.
\newblock Large language model (llm) for telecommunications: A comprehensive survey on principles, key techniques, and opportunities.
\newblock \emph{arXiv preprint arXiv:2405.10825}.

\end{thebibliography}
\bibliographystyle{acl_natbib}

\end{document}